\documentclass[10pt,twocolumn,letterpaper]{article}
\usepackage{amsmath}
\usepackage{float}
\usepackage[pagenumbers]{iccv} % To force page numbers, e.g. for an arXiv version
\usepackage{makecell}%
\usepackage{multirow}
\definecolor{iccvblue}{rgb}{0.21,0.49,0.74}
\usepackage[pagebackref,breaklinks,colorlinks,allcolors=iccvblue]{hyperref}
\usepackage{multirow}
\usepackage{multicol}
\usepackage{makecell}
\usepackage{bbding}

\def\paperID{9497} % *** Enter the Paper ID here
\def\confName{ICCV}
\def\confYear{2025}
\begin{document}
%%%%%%%%% TITLE - PLEASE UPDATE
\title{WonderTurbo: Generating Interactive 3D World in 0.72 Seconds}

%Interactive 3D Scene Generation from a Single Image

% Generating Interactive 3D World in 1 Second from a Single Image
%%%%%%%%% AUTHORS - PLEASE UPDATE
\author{
~~~~~~Chaojun Ni\footnotemark[1]~\textsuperscript{\rm 1,2}
~~~~~~Xiaofeng Wang\footnotemark[1]~\textsuperscript{\rm 1,3}
~~~~~~Zheng Zhu\footnotemark[1]~\textsuperscript{\rm 1}\textsuperscript{\Envelope} 
~~~~~~Weijie Wang\footnotemark[1]~\textsuperscript{\rm 1,4} 
~~~~~~Haoyun Li~\textsuperscript{\rm 1,3} \\
~~~~~~Guosheng Zhao~\textsuperscript{\rm 1,3}
~~~~~~Jie Li~\textsuperscript{\rm 1} 
~~~~~~Wenkang Qin\textsuperscript{\rm 1} 
~~~~~~Guan Huang\textsuperscript{\rm 1}
~~~~~~Wenjun Mei~\textsuperscript{\rm 2}\textsuperscript{\Envelope}
\\
\textsuperscript{\rm 1}GigaAI
~ ~ \textsuperscript{\rm 2}Peking University \\
~ ~ \textsuperscript{\rm 3}Institute of Automation, Chinese Academy of Sciences
~ ~ \textsuperscript{\rm 4}Zhejiang University
\\
\small{Project Page: \url{https://wonderturbo.github.io}}
}

\twocolumn[{%
\vspace{-1em}
\maketitle
\vspace{-2em}
\begin{center}

\centering
\setlength{\abovecaptionskip}{0.5em}
\resizebox{0.99\linewidth}{!}{
\includegraphics{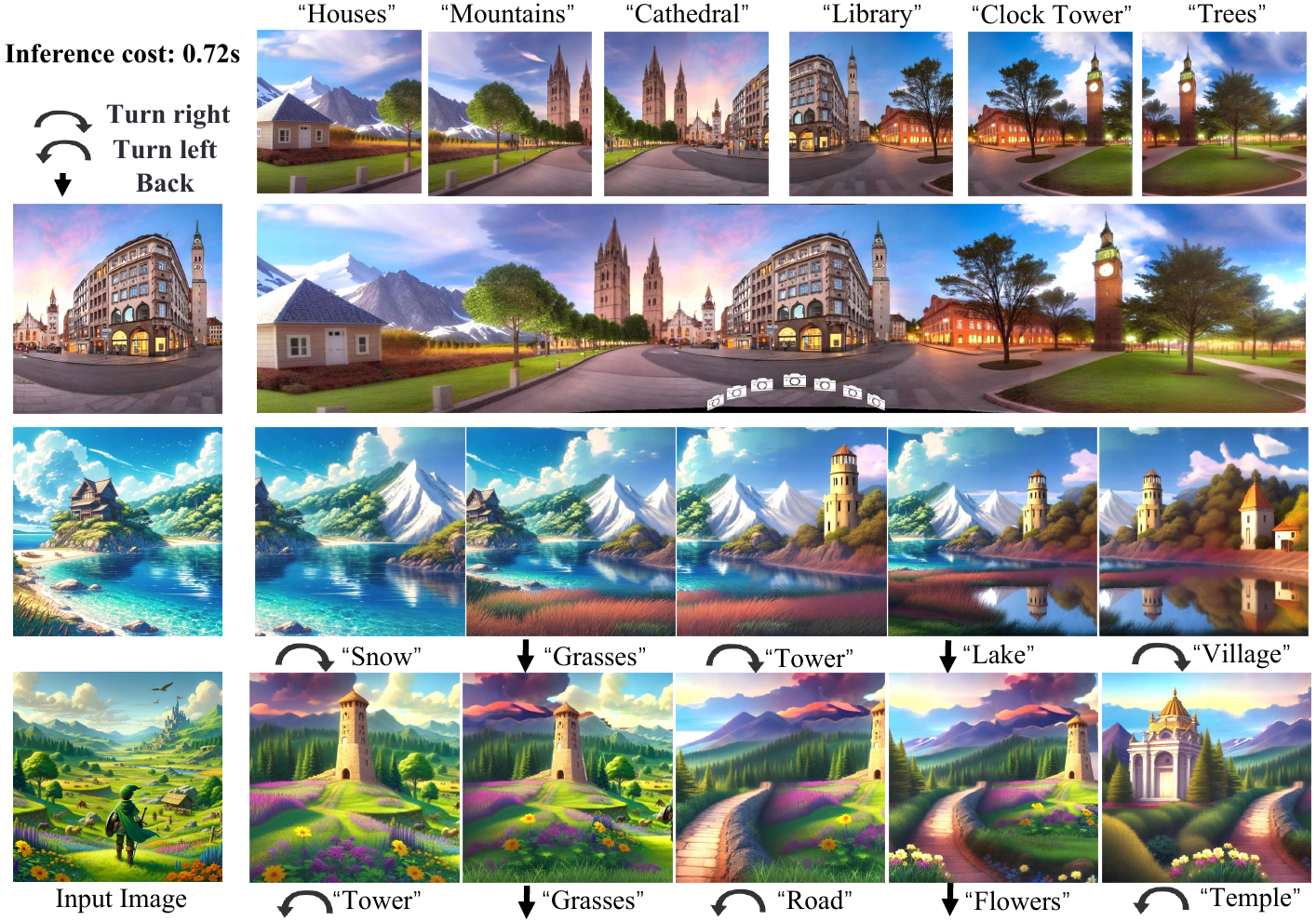}
}
\captionof{figure}{
Beginning with a single image, users can freely adjust the viewpoint and interactively control the generation of a 3D scene, each interaction requiring only 0.72 seconds.
}
\label{fig:main}
% \vspace{-0.5em}
\end{center}}]

\renewcommand{\thefootnote}{\fnsymbol{footnote}}
\footnotetext[1]{These authors contributed equally to this work.
\textsuperscript{\Envelope}Corresponding authors: Zheng Zhu, zhengzhu@ieee.org, Wenjun Mei, mei@pku.edu.cn.}

% #3d世界生成实时交互重要性 first challenge
\begin{abstract}
Interactive 3D generation is gaining momentum and capturing extensive attention for its potential to create immersive virtual experiences. However, a critical challenge in current 3D generation technologies lies in achieving real-time interactivity. To address this issue, we introduce WonderTurbo, the first real-time interactive 3D scene generation framework capable of generating novel perspectives of 3D scenes within 0.72 seconds. Specifically, WonderTurbo accelerates both geometric and appearance modeling in 3D scene generation. In terms of geometry, we propose StepSplat, an innovative method that constructs efficient 3D geometric representations through dynamic updates, each taking only 0.26 seconds. Additionally, we design QuickDepth, a lightweight depth completion module that provides consistent depth input for StepSplat, further enhancing geometric accuracy. For appearance modeling, we develop FastPaint, a 2-steps diffusion model tailored for instant inpainting, which focuses on maintaining spatial appearance consistency. Experimental results demonstrate that WonderTurbo achieves a remarkable 15$\times$ speedup compared to baseline methods, while preserving excellent spatial consistency and delivering high-quality output.

\end{abstract}

%补充下在线成成的定义
\section{Introduction}
The online generation of 3D from a single image \cite{wonderjourney,wonderworld}, which involves instant creation and updating of 3D scenes, has attracted significant attention. Compared with offline generation methods \cite{text2nerf,luciddreamer,genex,pano2room,text2room,recondreamer}, which generate a fixed 3D scene based on user texts and input images, online generation allows users to interactively create and edit 3D content, enhancing creation efficiency and flexibility. 
%For example, during film production, online 3D scene generation tools can be used to progressively adjust and create complex 3D environments, allowing full control over each step and immediate visualization of results after each modification, thereby significantly enhancing flexibility and customization. 

% However, despite these advantages, existing online 3D scene generation technologies still face significant challenges, largely constrained by processing time. This bottleneck mainly stems from the time-intensive processes of optimizing 3D scene representations and generating specific new content through diffusion-based models. Each expansion attempt requires more than 10 seconds to complete, which significantly impacts efficiency and user experience.

% Despite these advantages, existing online 3D scene generation techniques still face significant challenges, primarily due to time constraints. Current fastest 3D generation approach, WonderWorld, .....
% This bottleneck largely arises from the time-consuming processes of optimizing 3D scene representations and generating specific new content through diffusion-based models. For the former, 3D scene representation methods like 3DGS require training for specific scene representation. For the latter, diffusion inpainting models often require extensive sampling steps. Moreover, depth completion is highly time-consuming due to.... to maintain consistency.

Despite these advantages, existing online 3D scene generation methods \cite{wonderworld,wonderjourney}
 still face significant challenges, primarily due to the low inference efficiency. This efficiency bottleneck largely stems from the time-consuming processes of optimizing geometric details and generating or refining new view appearances.  Specifically, 3D scene representation methods like 3D Gaussian Splattings (3DGS)~\cite{3dgs} require iterative training to update new geometries, while appearance refinement is based on diffusion-based image inpainting models~\cite{sd} that require extensive inference steps, further increasing computational overhead. For example, the current fastest online 3D generation approach, WonderWorld \cite{wonderworld}, takes nearly 10 seconds to update a single 3D view, which falls short of real-time performance expectations. Although some works~\cite{stereo,shih20203d,layer,20193d,single,stereo,humandreamer} on generating novel views from a single image improve in speed, these methods only support generating views within small viewpoint changes.
%Wonderland \cite{wonderland} predicts 3DGS in a feed-forward manner by constructing 3D reconstruction models based on a video diffusion model's latent space.  This approach eliminates the dependence on iterative training to update new geometries but does not meet real-time speed requirements due to its reliance on a video diffusion-based method~\cite{svd,cogvideo}.

In this paper, we present \textit{WonderTurbo}, a novel framework designed for real-time interactive 3D scene generation. To address the critical challenges of inference efficiency, we optimize both geometric representation and appearance modeling.
For geometric optimization, we introduce \textit{StepSplat}, a scalable method that accelerates 3D scene expansion in 0.26 seconds. Unlike conventional 3DGS methods~\cite{3dgs,street,deformable,recondreamer,drivedreamer4d} that rely on iterative training to update 3D representations, \textit{StepSplat} leverages insights from recent feed-forward approaches~\cite{pixelsplat,mvsplat,depthsplat,wang2025freesplat}  to perform direct inference 3DGS. Moreover, \textit{StepSplat} extends the feed-forward paradigm to interactive 3D geometric representation, ensuring consistency across dynamic viewpoint changes. This is achieved through the maintenance of a feature memory module, which adaptively constructs cost volumes as the viewpoint changes. Meanwhile, to further enhance depth coherence, we incorporate a lightweight depth completion module, \textit{QuickDepth}, to provide a consistent depth prior for \textit{StepSplat} to construct the cost volume. On the appearance front, we propose \textit{FastPaint}, a highly efficient method for real-time appearance refinement. In contrast to traditional diffusion-based inpainting methods~\cite{sd}, which require dozens of inference steps to refine appearance modeling, \textit{FastPaint} achieves comparable results with only 2 inference steps while preserving spatial appearance consistency. 

We present the results of various camera setups, including a panoramic camera path and two casual walking camera paths as shown in Fig.~\ref{fig:main}. The results demonstrate that \textit{WonderTurbo} can accurately generate 3D scenes based on user-provided text while maintaining high consistency. Furthermore, our model achieves leading performance in CLIP-based metrics~\cite{clip,clipiqa} and user study win rates, while significantly boosting speed, achieving a 15$\times$ acceleration. 

The main contributions of this paper are summarized as follows:

\begin{itemize}

    \item We present \textit{WonderTurbo}, the first real-time (inference cost: 0.72 seconds) 3D scene generation method that allows users to interactively create diverse and cohesively connected scenes.
    \item For geometric efficiency optimization, the proposed \textit{StepSplat} extends feed-forward paradigm to interactive 3D geometric representation, accelerating the expansion of 3D scenes within 0.26 seconds. Besides, \textit{QuickDepth} is introduced to ensure depth consistency during viewpoint changes. For appearance modeling efficiency, we present \textit{FastPaint} for image inpainting with only 2-steps inference.
    \item We perform comprehensive experiments to validate that WonderTurbo, while achieving a 15$\times$ acceleration, surpasses other methods in generating high-quality 3D scenes, both in geometry and appearance.
    
\end{itemize}

% \begin{table}[t]
%     \centering
%     \caption{Time cost for generating a scene including modeling geometry and appearance on an H20 GPU. The methods are separated into offline and online categories.}
%     \footnotesize
%     \label{tbl:speed}
%     \setlength{\tabcolsep}{0.15cm}
%     \begin{tabular}{ccccc}
%     \toprule
%          &Method & Geometry (s) & Appearance (s)  & Total (s)\\
%         \midrule
%         \multirow{2}{*}{\rotatebox{90}{  Offline}} & LucidDreamer~\citep{luciddreamer} & 35.38 & 2.65 & 37.93    \\
%         & Text2Room~\citep{text2room} & 34.23 & 2.32 & 36.55 \\
%         & Pano2Room~\citep{pano2room} & 27.91 & 1.47 & 29.38 \\
%         & DreamScene360~\cite{dreamscene360} & 44.29 & 1.45 & 45.74 \\
%         \midrule
%         \multirow{2}{*}{\rotatebox{90}{Online}} & WonderJourney ~\cite{wonderjourney} & 78.12 & 1.45 & \ 79.57  \\
%         & WonderWorld~\cite{wonderworld} & 6.62 & 4.43 & 11.05 \\
%         & \textit{WonderTurbo} & \textbf{0.50} & \textbf{0.22} & \textbf{0.72}  \\
%     \bottomrule
%     \end{tabular}
% \end{table}

\section{Related Work}

%改成from
% 
\subsection{Offline 3D Scene Generation from Single Image}

Offline 3D scene generation from a single image ~\cite{text2nerf, genex, luciddreamer, pano2room, text2room, scenecraft, dreamscene, disentangled, gaudi} has been explored by various methods, which generally involve generating multiple views or panoramas of a scene and subsequently transforming them into a 3D representation. Approaches such as Text2Room~\cite{text2room} and LucidDreamer~\cite{luciddreamer} begin with a single input image and a user's textual description, generate multiple images of the scene, and then employ 3D optimization to refine the scene and create a more accurate and consistent 3D representation. Meanwhile, Wonderland \cite{wonderland} predicts 3DGS in a feed-forward manner by constructing 3D reconstruction models based on a video diffusion model's latent space. In contrast, methods such as GenEx~\cite{genex}, Pano2Room~\cite{pano2room}, and Dreamscene360~\cite{dreamscene360} synthesize coherent panoramas by leveraging pretrained text-to-panorama diffusion models~\cite{diffusion360,diffpano}, which are then elevated to 3D, ultimately producing explorable 3D worlds. However, these methods typically operate offline, preventing user interaction during the generation process. Furthermore, once the scene is generated, modifying or adjusting the layout or content becomes challenging.

%删除real time 区分两种 慢的 我们的 
\subsection{Online 3D Scene Generation from Single Image}
Some researches have focused on the online interactive generation of 3D scenes, which only requires an image to generate a new 3D scene and specify its content. WonderJourney~\cite{wonderjourney} employs an LLM~\cite{deepseek,qwen2,language,deepseek2} to generate scene descriptions, a text-driven pipeline for coherent 3D scene generation, and a vision-language model (VLM)~\cite{kuo2022f,bai2025qwen2} for result verification, while allowing users to adjust text and control the generation of new 3D scenes. However, this process takes several minutes, making it unsuitable for interactive use. WonderWorld~\cite{wonderworld} reduces reconstruction time using Fast LAyered Gaussian Surfels (FLAGS) and generates geometrically consistent scenes with guided diffusion-based depth estimation, but still requires about 10 seconds per scene. In contrast,~\textit{WonderTurbo} achieves the generation of diverse scenes within 0.72 seconds through accelerations both in geometric and appearance modeling, meeting the needs for real-time interaction.

%改成场景重建 分两部分 3dgs传统 feedforward的3dgs
\subsection{3D Scene Representations}
3DGS~\cite{3dgs} has attracted significant attention, for its high efficiency and photorealistic rendering. However, a major limitation of traditional 3DGS-based methods~\cite{deformable,streetgaussian,drivedreamer,recondreamer} is the need for per-scene optimization, which can be time-consuming. 2D Gaussian Splatting~\cite{2dgs} addresses this by projecting 3DGS onto a 2D plane, allowing faster rendering while maintaining geometric accuracy. Additionally, WonderWorld~\cite{wonderworld} introduces FLAGS, which employs a multi-layered representation and geometry-based initialization to reduce the need for per-scene optimization.

However, these methods still require significant computational time. Therefore, recent work~\cite{pixelsplat,mvsplat,wang2025freesplat,depthsplat,mvsplat360,splatt3r,ye2024no} has explored using feed-forward networks to predict 3D geometry from images. PixelSplat~\cite{pixelsplat} predicts 3DGS distributions by learning from paired images and MVSplat~\cite{mvsplat} leverages multi-view correspondence information through the construction of a cost volume. These methods are not suitable for generating interactive 3D scenes, especially for scenarios with gradually increasing views. Moreover, their reliance on unsupervised depth estimation leads to poor generalization.

Recent explorations~\cite{wang2025freesplat,depthsplat,crafting} have attempted to address these challenges. FreeSplat~\cite{wang2025freesplat} reconstructs long sequence inputs by constructing an adaptive cost volume between adjacent views and aggregating features through multi-scale structures. However, it lacks depth supervision and does not address the requirement for handling a gradually increasing number of views. DepthSplat~\cite{depthsplat} integrates monocular depth estimation priors~\cite{depthanything} into the feed-forward process, but its generalization capability remains limited, and it does not meet the needs of gradually increasing views.

\begin{figure*}[t]
\centering
\includegraphics[width=\textwidth]{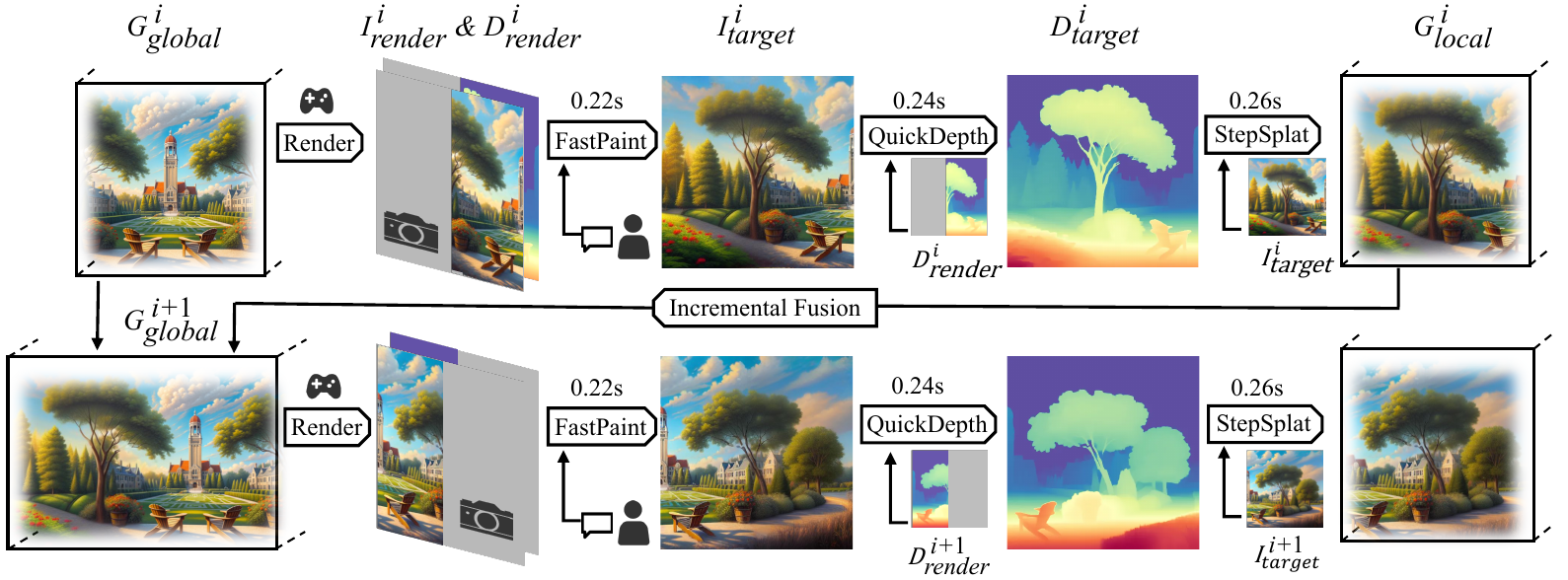}
\caption{The pipeline of \textit{WonderTurbo}. As the user moves the real-time rendering camera and inputs the text, the rendered image and depth map are then processed by \textit{FastPaint} and \textit{QuickDepth} to generate coherent geometry and appearance. Finally, \textit{StepSplat} performs incremental fusion based on the outputs of \textit{FastPaint} and \textit{QuickDepth}.}
\label{pipeline}
% \vspace{-1.5em}
\end{figure*}

%写成介绍 参考wonderworld
\section{Method}
\subsection{Overall Framework of WonderTurbo}
% Interactive 3D scene generation is constrained by time and computational cost. WonderWorld~\cite{wonderworld} accelerates geometry modeling through FLAGS to alleviate this issue, while real-time interaction still remains distant. In contrast, WonderTurbo achieves real-time interactive 3D scene generation by accelerating both geometry and appearance modeling. As shown in Fig.~\ref{pipeline}, we present the pipeline of \textit{WonderTurbo}. At $i$-th iteration, given a user-specified location, \textit{FastPaint} leverages the rendered image $I_{\text{g}}^{i}$ by the current 3D scene along with a user-provided textual description to craft a new scene appearance $I_{\text{t}}^{i}$. \textit{QuickDepth} is then employed to generate depth maps  $D_{\text{t}}^{i}$, where the input is the rendered depth map  $D_{\text{g}}^{i}$ and $I_{\text{t}}^{i}$, ensuring that the geometry of the newly generated scene aligns with the existing 3D scene. Finally, \textit{StepSplat} takes depth map $D_{\text{t}}^{i}$  and the new scene appearance $I_{\text{t}}^{i}$ as inputs, incrementally fusing ${\text{3DGS}}^{i}_{\text{local}}$ into the ${\text{3DGS}}^{i}_{\text{global}}$.

Interactive 3D scene generation~\cite{wonderworld,wonderjourney} is constrained by computational efficiency due to the time-consuming geometry and appearance modeling. WonderWorld~\cite{wonderworld} introduces FLAGS to accelerate geometric modeling. However, it still requires hundreds of iterations to optimize the geometry representation, and its appearance modeling relies on a pretrained diffusion model \cite{sd} that needs dozens of inference steps for inpainting. In contrast, \textit{WonderTurbo} achieves real-time interactive 3D scene generation by accelerating both geometry and appearance modeling.
Specifically, we propose \textit{StepSplat} for geometric modeling acceleration, which directly infers 3DGS in 0.26 seconds. Within this framework, \textit{QuickDepth} completes missing depth information in 0.24 seconds. For appearance modeling acceleration, we introduce \textit{FastPaint}, which completes image inpainting in 0.22 seconds.

We present the pipeline of \textit{WonderTurbo} in Fig.~\ref{pipeline}. At the $i$-th iteration, given a user-specified location, \textit{FastPaint} leverages the rendered image $I_{render}^i$ from the current 3D scene and a user-provided textual description to generate a new scene appearance $I_{target}^i$. Subsequently, \textit{QuickDepth} generates depth maps $D_{target}^i$ using the rendered depth map $D_{render}^i$ and the newly generated appearance $I_{target}^i$, ensuring that the geometry of the newly generated scene aligns with the existing 3D scene. Finally, StepSplat takes the depth map $D_{target}^i$ and the new scene appearance $I_{target}^i$ as inputs, incrementally fusing $G_{local}^i$ into the global representation $G_{global}^i$. In the following sections, we delve into the details of \textit{StepSplat}, \textit{QuickDepth} and \textit{FastPaint}.

\begin{figure}[t]
\centering
\setlength{\abovecaptionskip}{0.5em}
\includegraphics[width=0.45\textwidth]{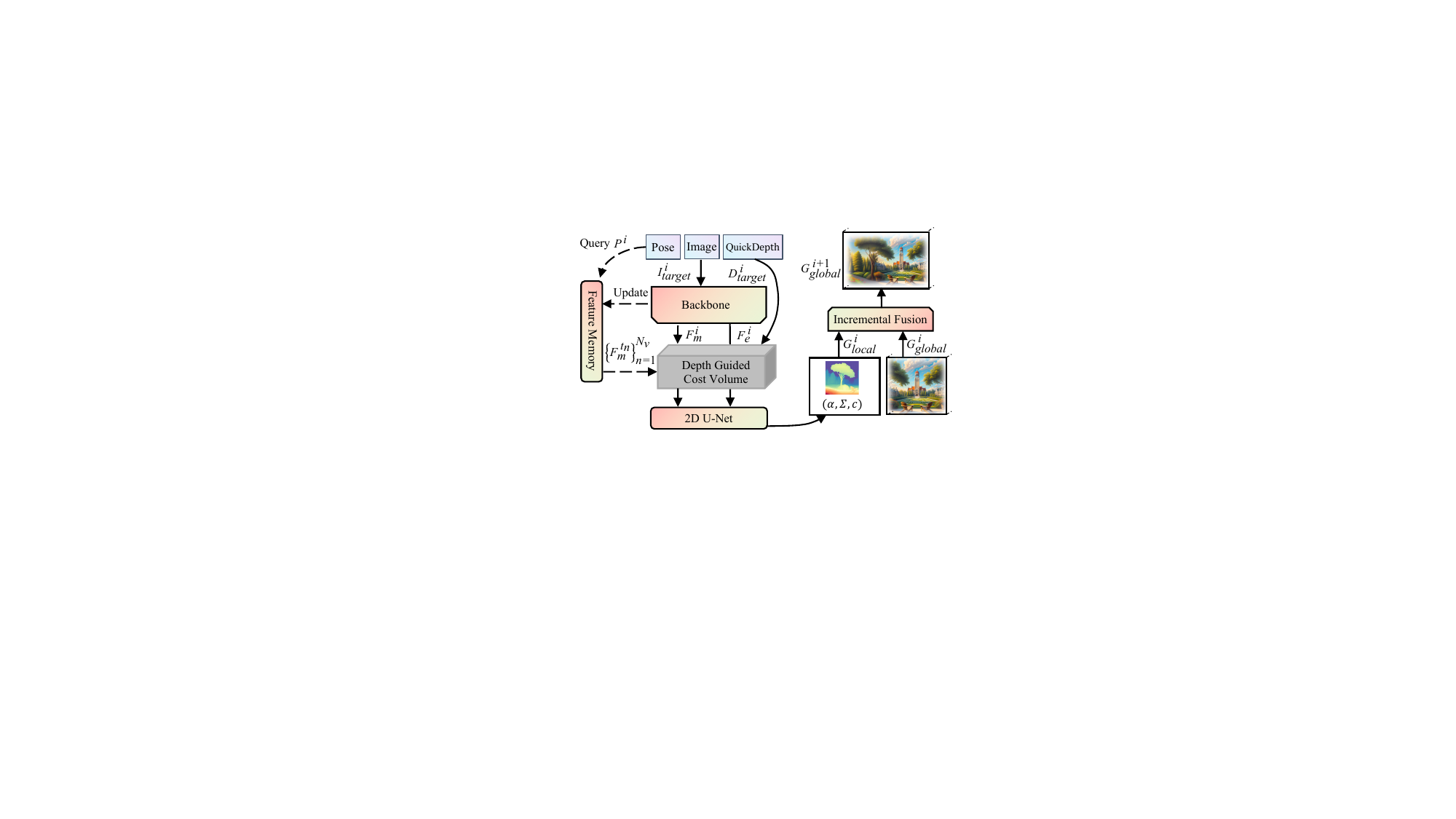}
\caption{The structure of \textit{StepSplat}.}
\label{stepsplat}
\vspace{-1.5em}
\end{figure}

\subsection{StepSplat}To accelerate the modeling of appearance, we introduce \textit{StepSplat}. As shown in Fig.~\ref{stepsplat}, \textit{StepSplat} takes as input the pose $ P^{i} $, image $ I_{target}^{i} $, and corresponding depth $ D_{target}^{i} $ from \textit{QuickDepth}, and first uses backbone to extract both the matching features $ F_{m}^{i} $ and image features $ F_{e}^{i} $. Then, it queries the feature memory for nearby views' matching features to construct cost volumes. Then, cost volumes is concatenated with $F_{e}^{i}$ to predict Gaussian parameters. Meanwhile, we leverage consistent input depth from \textit{QuickDepth} as a geometric priority to construct the cost volume, which ensures the accuracy of Gaussian centers. Finally, an incremental fusion strategy merges newly generated $G_{local}^{i}$ from the current view into the $G_{global}^{i}$, ensuring continuous and consistent 3D representation.

% ($F_{\text{e}}^{i} \in \mathbb{R}^{\frac{\text{H}}{\text{4}} \times \frac{\text{W}}{\text{4}} \times \text{C}}$)   ($F_{\text{m}}^{i} \in \mathbb{R}^{\frac{\text{H}}{\text{4}} \times \frac{\text{W}}{\text{4}} \times \text{C}}$)

\noindent{\textbf{Feature Memory.}} We introduce the feature memory to store the matching features of previous views, which are used to build the late cost volume. Given the input images $I^{i}_{target}$ and $P^{i}$, we first feed them into backbone to extract image features $F_{e}^{i}$ and matching features $F_{m}^{i}$. Then, we construct a tuple $(P^{i}, F_{m}^{i})$ that is updated into the feature memory. To accelerate inference, we adapt RepVGG~\cite{repvgg} as the backbone.
 
\noindent{\textbf{Depth Guided Cost Volume.}} For constructing the cost volume for the current view, we adaptively select $N_v$ neighboring views from the feature memory and use the input depth from \textit{QuickDepth} for the depth candidates of the cost volume. To achieve this, we first compute the distance between $P^i$ and all stored poses $\{P^{n}\}_{n=1}^{i-1}$ in the Feature Memory:  
\begin{equation}
d(P^n, P^i) = \|P^n - P^i\|_2,
\label{eq:distance}
\end{equation}
where $\|\cdot\|_2$ represents the L2 norm. From these distances, we select the $N_v$ closest poses to the current pose and extract their corresponding matching features from Memory as $\{(P^{t_n}, F^{t_n}_m)\}_{n=1}^{N_v}$, where each $t_n$ corresponds to one of the $N_\text{v}$ closest poses.

% where $\|\cdot\|_2$ represents the L2 norm and $P^t$ denotes the pose of any stored view $t$. Subsequently, we select the $N$ closest views to the current view $P^t$ as the neighboring view set $\mathcal{N}_t$. Then, extract the corresponding poses and matching features from Memory:
% \begin{equation}
% \{(P^{t_n}, F^{t_n}_m)\}_{n=1}^N = \{(P^i, F^i_m) \mid i \in \mathcal{N}_t\}
% \label{eq:feature_extraction}
% \end{equation}

For ensuring consistent 3D representation, inspired by Multi-View Stereo~\cite{shao2022smudlp,disentangling,crafting}, we use \( D_{target}^{i} \) to guide the construction of the cost volume. Specifically, $N_d$ depth candidates $\{d_s\}_{s=1}^{N_d}$ are uniformly sampled from \( R \) as:
\begin{equation}
R = \{ d \mid (1 - a) \cdot D_{target}^i \leq d \leq (1 + a) \cdot D_{target}^i \},
\end{equation}
where \( a \) is the offset value used to adjust the depth candidate range. Then, each neighboring view's matching feature \( {F}^{t_n}_m \) is warped to the candidate depth $d_s$ planes of the current view using the plane-sweep stereo algorithm~\cite{collins1996space}, the feature warping is formulated as:  
\begin{equation}
{F}^{t_n \to i}_{d_s} = \mathcal{W}({F}^{t_n}_m, {P}^i, {P}^{t_n}, d_s),
\label{eq:warp_multi}
\end{equation}
where $\mathcal{W}$ denotes the differentiable warping operation. We then compute the normalized dot-product correlation between the current view's feature \( {F}_m^i \) and each warped neighboring feature \({F}^{i_n \to i}_{d_s} \) , and  average the correlation maps from all neighboring views:
% \begin{equation}
% {S}^i = \left[ \frac{1}{N} \sum_{n=1}^N \frac{{F}_m^i \cdot {F}^{i_n \to i}_{d_m}}{\sqrt{C}} \right]_{m=1}^{N_{d}} ,
% \end{equation}
\begin{equation}
    S^i = \left[ \frac{1}{N_v} \sum_{n=1}^{N_v} F_m^i \cdot F^{t_n \to i}_{d_s} \right]_{s=1}^{N_d},
\end{equation}
where \( N_d \) is the number of candidate depths, and the correlation maps from each depth are stacked to form the cost volume ${S}^i$. Meanwhile, an additional 2D U-Net~\cite{unet} is used to further refine and upsample the cost volume. We normalize the cost volume \( S^i \) and perform a weighted average of all depth candidates to obtain the predicted depth map \( \hat{d} \):

\begin{equation}
\hat{d} = \text{softmax}(S^{i}) \cdot d.
\end{equation}

% Then, we first normalize the cost volume $S^i$ and then perform a weighted average of all depth candidates to get the prtedict depth map \hat{d}
% \begin{equation}
% \hat{d} = \text{softmax}(S^{i}) \cdot d ,
% \end{equation}

% Then, We use the softmax operation to obtain depth predictions. Specifically, for each pixel \(x, y) \), we compute the depth candidate probabilities using the softmax operation on the corresponding cost volume \( S^{i}_{x, y}(d_m) \). The final depth estimation \( \hat{d}_{x,y} \) for each pixel is then obtained by performing a weighted average over all depth candidates:

% \begin{equation}
% \hat{d}_{x,y} = \sum_{m=1}^{D} \text{softmax}(S^{i}_m)) \cdot d_m
% \end{equation}

After obtaining the depth predictions, the depth values are unprojected as the center of the 3DGS. The cost volume and image feature are then decoded to obtain other Gaussian parameters, similar to MVSplat~\cite{mvsplat}.

% \noindent{\textbf{Depth Estimation and Gaussian Parameters Prediction.}} We use the softmax operation to obtain depth predictions. Specifically, for each pixel \( (x, y) \), we compute the depth candidate probabilities using the softmax operation on the corresponding cost volume \( C^{i}_{x, y}(d_m) \). The final depth estimation \( \hat{d}_{x,y} \) for each pixel is then obtained by performing a weighted average over all depth candidates:

% \begin{equation}
% \hat{d}_{x,y} = \sum_{m=1}^{D} \text{softmax}(C^{i}_{x, y}(d_m)) \cdot d_m
% \end{equation}

% For opacity $\alpha$, we compute the matching distribution via $\mathrm{softmax}(C_t^i)$, take its maximum confidence value, and predict through a convolutional layer. Similarly, another two convolutional layers predict covariance $\Sigma$ and color $c$, using the refined cost volume and feature $F^i_{e}$ as inputs.

% For opacity $\alpha$, we first compute the matching distribution via $\mathrm{softmax}(C_t^i)$, then take its maximum confidence value, and finally predict through a two-stage convolutional layer. Similarly, we use two convolutional layers to predict the covariance $\Sigma$ and color $c$, with these layers taking the refined cost volume and feature $F_{ie}$ as inputs.

\noindent{\textbf{Incremental Fusion.}} To reduce the redundancy of Gaussians, we achieve the incremental fusion by updating $G^{i}_{local}$ to $G^{i+1}_{global}$ using a depth constraint. Specifically, given $G^{i}_{local}$ with a 2D coordinate $[x_{local}, y_{local}]$ and depth $d_{local}$, we project all global Gaussians $\{\mu_{global}^j\}_{j=1}^K$ from $G^{i}_{global}$ onto the current pixel coordinate system using the camera projection matrix $\mathbf{P}^i$:
\begin{equation}
    \left[x_j^g, y_j^g, d_j^g\right]^\top = \mathbf{P}^i \mu_{global}^j,
\end{equation}
 We then construct a candidate set of global Gaussians projected to the same discrete pixel location via:
\begin{equation}
    \mathcal{S}_{global} = \left\{j \, \left| \, \lfloor x_j^g \rfloor = x_{local} \, \land \, \lfloor y_j^g \rfloor = y_{local} \right. \right\}.
\end{equation}
To enforce geometric continuity, we prune conflicting Gaussians from $\mathcal{S}_{global}$ that violate the depth consistency constraint. The Gaussians to be pruned are those in $\mathcal{C}$, defined as:
\begin{equation}
    \mathcal{C} = \left\{j \in \mathcal{S}_{global} \, \left| \, |d_{local} - d_j^g| < \delta \cdot d_{local} \right. \right\},
\end{equation}
where $\delta$ controls the depth tolerance. The global model is then updated by selectively merging the valid local Gaussians, which are those not included in $\mathcal{C}$, into the existing global model, as shown by:
\begin{equation}
G^{i+1}_{global} \leftarrow G^{i}_{global} \cup \left( \text{G}^i_{local} \setminus \mathcal{C}\right).
\end{equation}

%数据集单独放到最后。
\noindent{\textbf{Training of StepSplat.}} Traditional 3DGS feed-forward methods~\cite{mvsplat,pixelsplat,depthlab,wang2025freesplat} struggle to meet the demands of interactive 3D scene generation. This is partly due to the limited diversity of datasets, which focus on specific scenes such as autonomous driving~\cite{nuscenes,waymo} or indoor environments~\cite{scannet,replica}. Additionally, there is a significant gap between the viewpoint variations in these datasets and the requirements of interactive 3D scene generation. 
To address this challenge, we creat a dataset utilizing 3D generation models~\cite{wonderworld,wonderjourney,pano2room,text2room,luciddreamer} with simulated viewpoint changes for the purpose of training StepSplat, which is detailed in Section.~\ref{sec:Dataset}. During training, we randomly select an image sequence and feed these images into the model one by one to generate a global Gaussian representation. This representation is then used to render images from novel viewpoints, with the RGB images serving as supervision.

% To train  \textit{StepSplat}, we build a comprehensive dataset named OmniScene, which includes both indoor and outdoor datasets, and generates additional stylistically varied datasets using a 3D scene generation model. Specifically, we create 3D scenes that include realistic and synthetic images by xx. These 3D scenes are then utilized to render 2D images and their corresponding depth maps for various trajectories, with more details provided in the supplementary material. During training, we sample random image sequences and feed them into  \textit{StepSplat} one by one. We use ground truth RGB images as supervision.

\begin{figure}[h]
\centering
\setlength{\abovecaptionskip}{0.5em}
\includegraphics[width=0.45\textwidth]{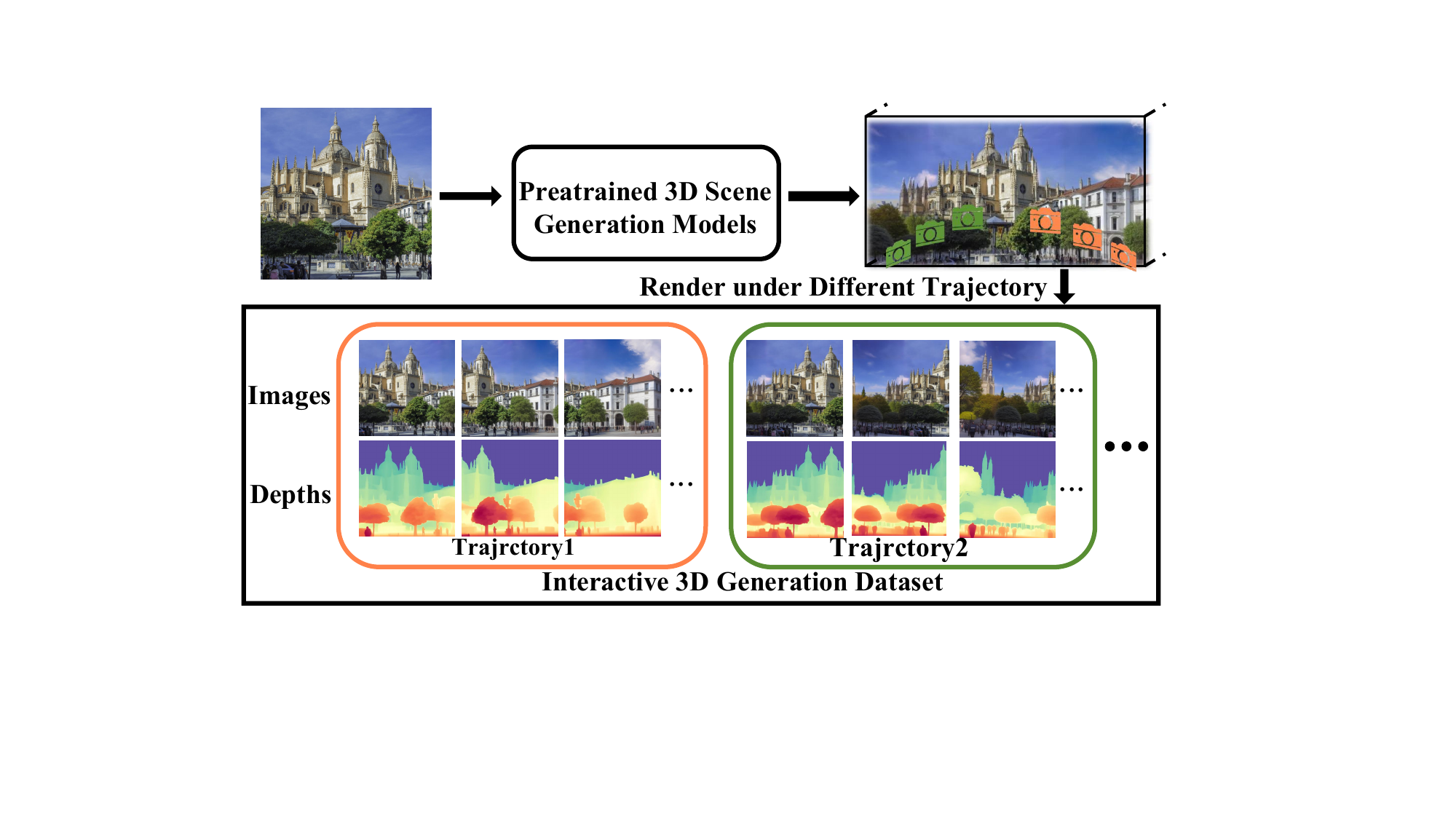}
\caption{The process of constructing the interactive 3D generation dataset.}
\label{dataset}
\vspace{-1.5em}
\end{figure}

\subsection{QuickDepth}
% We train depth completion network using our dataset, OmniScene. Unlike networks trained exclusively on indoor or outdoor scenes, utilizing data generated by 3D generative models provides several advantages. Firstly, the diversity of the generated data significantly enhances the network's generalization capabilities. Secondly, the ability to prescribe data collection trajectories ensures that the training data closely aligns with applications of Wonderturbo.
In the field of depth completion, existing methods~\cite{infusion,fcfr,depthlab,depthanything,midas} have made notable advancements. However, these methods are generally designed for sparse depth completion and face challenges in completing depth for regions that lack any depth information, a critical requirement for interactive 3D scene generation. To address this, WonderWorld~\cite{wonderworld} introduces a training-free guided depth diffusion method, but it requires over 3 seconds per depth map. Invisible Stitch~\cite{invisible} trains a depth completion model through teacher distillation and self-training due to the lack of ground-truth data. However, its training data are limited, leading to a decline in performance in some scenarios. We introduce \textit{QuickDepth}, a lightweight depth completion model trained on our constructed dataset for interactive 3D scene generation, with strong generalization capabilities, performing well across a wide range of scenarios.

To adapt \textit{QuickDepth} to interactive 3D scene generation, we construct a dataset comprising diverse scenes, including indoor and outdoor environments, and scenes from comics and artworks (detailed in Section~\ref{sec:Dataset}). Instead of using random masks or projections to simulate the mask in interactive 3D scene generation~\cite{invisible}, we design a series of camera trajectories that are more aligned with interactive 3D scene generation. Specifically, we design camera poses $\{T_1, \ldots, T_n\}$ and obtain frames $\{I_1, \ldots, I_n\}$ along with their corresponding depth maps $\{D_1, \ldots, D_n\}$ from the dataset. We then utilize the geometric relationships between adjacent frames. More precisely, for each frame $I_j$ ($j \in [1, n]$), we project the depth map of the previous frame $D_{j-1}$ into the coordinate system of $I_j$ using the relative pose $T_{j-1 \to j}$. This process produces incomplete depth $D'_{j-1 \to j}$ and binary validity mask $M_{j-1 \to j}$, where invalid pixels indicate regions that require depth completion.

% Instead of using random masks, we leverage the geometric relationships between consecutive frames to generate masks. Given consecutive frames $\{I_1, \ldots, I_t\}$, along with their corresponding depth maps $\{D_1, \ldots, D_t\}$ and camera poses $\{T_1, \ldots, T_t\}$ from the dataset, for each frame $I_j$ ($j \in [1, t]$), we project all other frames' depth maps $\{D_k\}_{k \neq j}$ to the coordinate system of $I_j$ using the relative poses $\{T_{k \to j}\}_{k \neq j}$. This process produces incomplete depth estimates $\{D'_{k \to j}\}$ and binary validity masks $\{m_{k \to j}\}$, where invalid pixels indicate regions requiring depth completion.

During training, we construct inputs by either masking the target frame's ground truth depth $ D_{j} $ entirely or selecting a warped depth-mask pair $ (D'_{j-1 \to j}, M_{j-1 \to j}) $. Meanwhile, \textit{QuickDepth} is initialized with a light pre-trained depth estimation model~\cite{bhat2023zoedepth} and takes as input the target frame's RGB image, the incomplete depth map and the binary mask. It then predicts the full depth, where the prediction is supervised by the target's ground truth depth through a $ L_1 $ loss. 

\subsection{FastPaint}
% Existing 3D Scene Generation from a Single Image methods based on the Stable Diffusion Inpaint model face significant challenges in real-time generation. Traditional diffusion models typically require multiple inference steps to complete generation, which incurs substantial time costs. Moreover, the generated appearances often fail to align with user expectations, necessitating post-processing checks using vision large language models (VLMs) to ensure usability. To address these issues, we propose FastPaint , an efficient inpainting method that reduces the diffusion process to just 2 inference steps , enabling faster generation while preserving high-quality 3D appearances.
In 3D scene generation, image inpainting~\cite{shadow,continuous,autoregressive,your} is crucial for modeling 3D appearance. Some methods, like Pano2Room~\cite{pano2room}, generate panoramic images from a single input but struggle to place content at user-specified locations. Others, such as WonderJourney~\cite{wonderjourney} and WonderWorld~\cite{wonderworld}, use Stable Diffusion-based fine-tuned inpainting models~\cite{sd}. However, during fine-tuning, the inpainting regions differ significantly from those in 3D scene generation, requiring a separate model to verify generated content for each instance~\cite{wonderjourney}. Moreover, these diffusion models need multiple inference steps. To address this, we propose \textit{FastPaint}, which reduces inference to 2 steps and enhances pretrained models’ inpainting capability through distillation and fine-tuning, making them suitable for interactive 3D generation.

Specifically, our approach synergistically leverages the advantages of ODE trajectory preservation and reformulation~\cite{hypersd} to perform knowledge distillation~\cite{knowledge} on the pretrained model~\cite{sd}, reducing the number of inference steps required while maintaining the quality of appearance modeling. To address the issue where inpainting regions in 3D scene generation differ from those in fine-tuning,  a dataset is constructed for training \textit{FastPaint}. Especially, we designed camera poses to simulate the interactive 3D generation process. By obtaining depth maps and images and using projections to acquire masks, we ensure the dataset aligns with the specific requirements of the inpainting task in this context. The construction of this dataset shares similarities with the methodology used in \textit{StepSplat} and \textit{QuickDepth}, particularly in simulating camera trajectories to help the models adapt to interactive 3D scene generation.

% particularly in simulating camera trajectories to enable the models to better adapt to the demands of interactive 3D scene generation.

\subsection{Interactive 3D Generation Dataset}
\label{sec:Dataset}
% 数据集的统计特性
Interactive 3D generation from a single image allows for a wide variety of images with different styles to be used as inputs. However, real-world data is often limited to a few specific scenes, mainly in autonomous driving~\cite{waymo,nuscenes} or indoor environments~\cite{replica,scannet}. This limitation results in poor generalization for current 3D generation methods~\cite{pano2room,dreamscene360,luciddreamer}. Meanwhile, some methods~\cite{wonderjourney,wonderworld} directly use pretrained models to construct the pipeline, which may not be tailored for 3D interactive scene generation, resulting in the need for a Vision-Language Model (VLM)~\cite{VLM-guided,kuo2022f,bai2025qwen2} to verify if the generated content matches the scene style or textual requirements. 

To overcome this limitation, we build a dataset based on current 3D scene generation methods and train all our modules using this dataset. Specifically, we employ multiple 3D scene generation methods~\cite{wonderjourney,wonderworld,text2room,luciddreamer,pano2room,dreamscene360,realmdreamer,zhang20243d} to create 3D scenes that each method excels at. Meanwhile, a VLM model~\cite{wonderjourney} is used to verify whether the generated data conforms to the defined scene. Finally, the dataset contains over 6 million frames rendered through simulated interactive trajectories, including rotational paths, linear movements, and hybrid trajectories. The dataset primarily covers four categories: indoor environments (32\%), urban landscapes (28\%), natural terrains (25\%), and stylized artistic scenes (15\%). When training \textit{StepSplat}, we impose certain restrictions on the distances between adjacent frames as inputs for StepSplat. This is to avoid using frames that are too close together, ensuring better alignment with the practical application of 3D interactive generation. For \textit{FastPaint} and \textit{QuickDepth}, the depth of two adjacent frames is used to project and obtain the corresponding mask.

% To ensure mask diversity, the distance between adjacent frames is varied when setting the paths.

% It ensures that we can adapt the pretrained models for interactive 3D scene generation and enhance their generalization capability.

\section{Experiments}
In this section, we present our experimental setup, which includes implementation details and evaluation metrics. Subsequently, both quantitative and qualitative results are provided to demonstrate the superiority of \textit{WonderTurbo} in performance and efficiency. Finally, we conduct ablation studies to validate the effectiveness of each module.

\subsection{Experiment Setup}
\noindent
\textbf{Baselines.} In our comparative analysis, we select representative 3D generation methods, encompassing both offline and online approaches~\cite{luciddreamer,dreamscene360,pano2room,text2room,wonderworld,wonderjourney}. The offline methods include LucidDreamer~\cite{luciddreamer} and Text2Room~\cite{text2room}, which generate 3D scenes by producing multi-view images of a scene, as well as Pano2Room~\cite{pano2room} and DreamScene360~\cite{dreamscene360}, which directly generate panoramic images that are then elevated to 3D. For online 3D scene generation, we evaluate WonderJourney~\cite{wonderjourney} and WonderWorld~\cite{wonderworld}. All comparisons are conducted using the official codebases provided by these methods.

\noindent
\textbf{Evaluation Metrics.}
To evaluate the quality of 3D scene generation, following WonderWorld \cite{wonderworld}, we used CLIP \cite{clip} scores (CS), CLIP \cite{clip} consistency (CC), CLIP-IQA+ \cite{clipiqa} (CIQA), Q-Align \cite{qalin}, and CLIP aesthetic \cite{clip} scores (CA) as metrics. Additionally, a user study is conducted to gather subjective feedback on visual quality. More details are provided in the supplementary materials.

\noindent
\textbf{Implementation Details.} To ensure comprehensive evaluation, we use input images from LucidDreamer \cite{luciddreamer}, WonderJourney \cite{wonderjourney}, and WonderWorld \cite{wonderworld}. We generate 8 scenes for each of 4 test cases, totaling 32 scenes. Evaluations are conducted with a fixed panoramic camera. For efficiency, we compare the time taken to generate scenes of the same size within the camera's view. More details are in the supplementary materials.
\subsection{Main Results}

\begin{table}[t]
    \centering
    \caption{Time comparison of scene generation, including geometry and appearance modeling, on an H20 GPU for offline and online methods. We compare the time taken to generate scenes of the same size within the camera's view.}
     \vspace{-1.5mm}
    \footnotesize
    \label{tbl:speed}
        \resizebox{\columnwidth}{!}{
    \setlength{\tabcolsep}{0.15cm}
    \begin{tabular}{ccccc}
    \toprule
         & Method & Geometry (s) & Appearance (s)  & Total (s)\\
        \midrule
        \multirow{4}{*}{{\rotatebox{90}{Offline}}} % 改为跨4行并垂直居中
        & LucidDreamer~\citep{luciddreamer} & 35.38 & 8.32 & 43.70    \\
        & Text2Room~\citep{text2room} & 34.23 & 7.32 & 41.55 \\
        & Pano2Room~\citep{pano2room} & 27.91 & 1.47 & 29.38 \\
        & DreamScene360~\cite{dreamscene360} & 44.29 & 1.45 & 45.74 \\
        \midrule
        \multirow{3}{*}{{\rotatebox{90}{Online}}} % 改为跨3行并垂直居中
        & WonderJourney ~\cite{wonderjourney} & 78.12 & 1.45 & 79.57  \\
        & WonderWorld~\cite{wonderworld} & 6.62 & 4.43 & 11.05 \\
        & \textit{WonderTurbo} & \textbf{0.50} & \textbf{0.22} & \textbf{0.72}  \\
    \bottomrule
    \end{tabular}
    }
     \vspace{-3.5mm}
\end{table}

\begin{figure*}[!t]
\centering
\setlength{\abovecaptionskip}{0.2em}
\includegraphics[width=1.\textwidth]{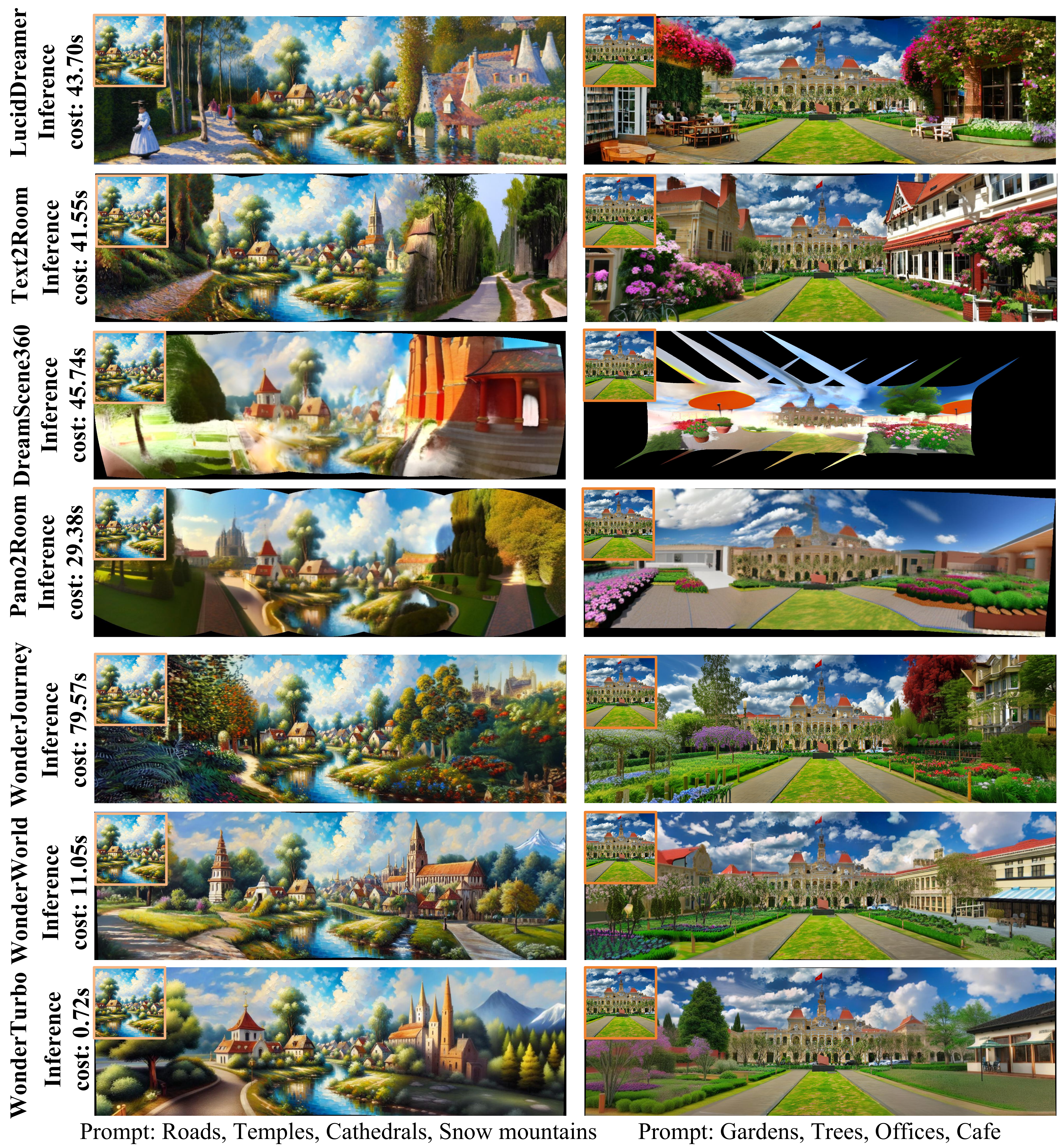}
\caption{Qualitative comparisons of using a fixed panoramic camera path.}
\label{fig:exp}
\vspace{-1.5em}
\end{figure*}

\noindent
\textbf{Generation Speed.} The time cost is crucial for interactive 3D scene generation. However, despite utilizing FLAGS for acceleration, WonderWorld \cite{wonderworld}, the fastest among the compared methods, still requires more than 10 seconds to generate a scene as shown in Tab.~\ref{tbl:speed}. LucidDreamer~\cite{luciddreamer} and Text2Room~\cite{text2room} involve generating multiple views for each new scene, substantially increasing the time devoted to modeling appearance. While Pano2Room~\cite{pano2room} and DreamScene360~\cite{dreamscene360} do not generate multiple perspectives, the inherent delays in panoramic image generation and the necessity for per-scene optimization significantly constrain their efficiency. Notably, \textit{WonderTurbo} excels in both geometry and appearance modeling, accelerating  overall time by 15$\times$.
% The time cost is crucial for interactive 3D scene generation. WonderWorld\cite{wonderworld} stands out as the fastest method among the baselines, yet it still requires more than 10 seconds to generate a scene. Especially while employing FLAGS to accelerate, it still necessitates 2.9s to optimize geometry modeling. Meanwhile, LucidDreamer and text2room need to generate and refine multiple views for a new scene, resulting in mush time comsumt in modeling appearance. Pano2room and DreamScene360 does not need to create multiple perspectives, but the time required for the panoramic image generation process itself, coupled with the requirement of per-scene optimization to achieve geometric consistency, limit overall efficiency. However, WonderTurbo 在geometry and appearance modeling上面都有领先，尤其在geometry上相较于最快的方法加速了20倍，总时间上领先最快的方法10倍。

\noindent
\textbf{Quantitative Results.}
In Tab.~\ref{tbl:metrics}, we compare \textit{WonderTurbo} with various 3D scene generation methods~\cite{luciddreamer,text2room,pano2room,dreamscene360,wonderworld,wonderjourney}. The experimental results show that online 3D scene generation methods outperform offline approaches by better meeting user textual requirements, achieving higher CLIP scores and improved CLIP consistency. WonderWorld~\cite{wonderworld} surpasses all other baseline methods, leading across all metrics. However, even with a 15$\times$ acceleration, \textit{WonderTurbo} maintains competitive performance across all metrics compared to WonderWorld~\cite{wonderworld}. Additionally, since~\textit{WonderTurbo} is fine-tuned specifically for interactive 3D generation tasks, improvements are observed in CLIP scores, CLIP consistency, CLIP-IQA+ and CLIP aesthetic.

\noindent
\textbf{User Study.}
Additionally, we conduct a user study to evaluate the quality of 3D scenes generated by various methods. As shown in Tab.~\ref{tbl:user}, the results indicate that \textit{WonderTurbo} achieves comparable performance to WonderWorld~\cite{wonderworld} with a lower scene generation time cost and significantly outperforms all other methods in terms of user preference.

\begin{table}[t]
    \centering
    \caption{Evaluation of novel view renderings for offline and online methods}
    \vspace{-1.5mm}
    \footnotesize
    \label{tbl:metrics}
        \resizebox{\columnwidth}{!}{
    \setlength{\tabcolsep}{0.15cm}
    \begin{tabular}{clccccc}
    \toprule
     & Method& CS$\uparrow$ & CC$\uparrow$ & CIQA$\uparrow$ & Q-Align$\uparrow$ & CA$\uparrow$ \\
    \midrule
    \multirow{4}{*}{{\rotatebox{90}{Offline}}}
    & LucidDreamer~\citep{luciddreamer} & 27.72 & 0.9213 & 0.6023 & 3.5439 & 6.8231    \\
    & Text2Room~\citep{text2room} & 24.50 & 0.9035 & 0.4910 & 2.6732 & 6.5324 \\
    & Pano2Room~\citep{pano2room} & 25.67 & 0.8652 & 0.3534 & 2.1342 & 5.0367 \\
    & DreamScene360~\cite{dreamscene360} & 24.50 & 0.8435 & 4.6973 & 2.4620 & 6.9846 \\
    \midrule
    \multirow{3}{*}{{\rotatebox{90}{Online}}}
    & WonderJourney~\cite{wonderjourney} & 27.63 & 0.9652 & 0.4753 & 3.5276 & 7.0134  \\
    & WonderWorld~\citep{wonderworld} & 28.14 & 0.9654 & 0.6764 & \textbf{3.7823} & 7.2121 \\
    & \textit{WonderTurbo} & \textbf{28.65} & \textbf{0.9732} & \textbf{0.6812} & 3.7253 & \textbf{7.3243} \\
    \bottomrule
    \end{tabular}
    }
    \vspace{-2.5mm}
\end{table}

\begin{table}[t]   
    \centering
    \caption{Comparing the win rates of \textit{WonderTurbo} in rendering
novel views.}
    \vspace{-1.5mm}
    \footnotesize
    \label{tbl:user}
        \resizebox{\columnwidth}{!}{
    \setlength{\tabcolsep}{0.1cm}
    \begin{tabular}{cccc}
    \toprule
    Method  &  Win Rate   & Method  & Win Rate  \\
    \midrule
    vs. LucidDreamer~\cite{luciddreamer}  & 96.32\% & vs. Text2Room~\cite{text2room}   & 98.47\% \\
    vs. Pano2Room~\cite{pano2room}  & 94.26\% & vs. DreamScene360~\cite{dreamscene360}  & 96.23\% \\
    vs. WonderJourney~\cite{wonderjourney}  & 96.54\% &vs. WonderWorld~\cite{wonderworld}  & 69.43\% \\
    
    \bottomrule
    \end{tabular}
    }
    \vspace{-5.5mm}
\end{table}

% \textbf{Qualitative Results.} 

% In Fig.~\ref{fig:exp}, we present a qualitative comparison between \textit{WonderTurbo} and baseline methods using identical input images, camera paths, and text prompts. The results demonstrate that 
% \textit{WonderTurbo} maintains competitive quality in scene generation while significantly reducing the generation time. DreamScene360~\cite{dreamscene360} and Pamo2room~\cite{pano2room} exhibit noticeable geometric distortions and lack appealing aesthetics due to inadequate generalization. LucidDreamer~\cite{luciddreamer} and Text2room~\cite{text2room} struggle to generate reasonable appearances in the appropriate locations, with some content specified in the prompt failing to materialize. WonderJourney~\cite{wonderjourney} displays significant distortions at the boundaries, and the generated content diverges from the specifications of the prompt. However, the results from \textit{WonderTurbo} and WonderWorld~\cite{wonderworld} are closely matched, both exhibiting robust performaunsupervised depth estimation
\noindent
\textbf{Qualitative Results.} As shown in Fig. \ref{fig:exp}, we present a qualitative comparison between \textit{WonderTurbo} and several baseline methods using the same settings. Notably, \textit{WonderTurbo} delivers competitive scene generation quality while significantly reducing generation time. In contrast, DreamScene360 \cite{dreamscene360} and Pamo2Room~\cite{pano2room} struggle with noticeable geometric distortions and lack aesthetic appeal due to limited generalization capabilities. Meanwhile, LucidDreamer~\cite{luciddreamer} and Text2Room~\cite{text2room} fail to place content correctly, with some prompt details not materializing.  The results from \textit{WonderTurbo} and WonderWorld~\cite{wonderworld} are closely matched, both demonstrating strong performance.

%In WonderJourney~\cite{wonderjourney}, significant distortions appear at the boundaries, especially on the left side of the first column of images and the generated content diverges from the prompt's specifications.

\begin{table}[t]
    \centering
    \caption{Ablation study results on different geometry models.}
    \vspace{-1.5mm}
    \footnotesize
    \label{tbl:geometry}
    \resizebox{\columnwidth}{!}{
    \setlength{\tabcolsep}{0.1cm}
    \begin{tabular}{lccccc}
    \toprule
         & CS$\uparrow$ & CC$\uparrow$ & CIQA$\uparrow$ &Q-Align$\uparrow$ & CA$\uparrow$ \\
        \midrule
        \textit{WonderTurbo} w/ FreeSplat & 27.65 & 0.9542 & 0.6460 &  3.1543 &     6.6235\\
        \textit{WonderTurbo} w/ DepthSplat & 27.32 & 0.9675 & 0.6620 & 3.2145 & 6.7432\\
        \textit{WonderTurbo} w/ \textit{StepSplat} & \textbf{28.65} &  \textbf{0.9732} & \textbf{0.6812} & \textbf{3.7253} & \textbf{7.3243} \\
    \bottomrule
    \end{tabular}
    }
    \vspace{-2.5mm}
\end{table}
\vspace{-1.5mm}

% \begin{table}[t]
%     \centering
%     \caption{Ablation study results on novel view renderings.}
%     \footnotesize
%     \label{tbl:ablation}
%     \setlength{\tabcolsep}{0.08cm}
%     \begin{tabular}{lccccc}
%     \toprule
%          & CS$\uparrow$ & CC$\uparrow$ & CIQA$\uparrow$ &Q-Align$\uparrow$ & CA$\uparrow$ \\
%         \midrule
%         Ours w/o depth guided & 26.72 & 0.9532 & 0.6359 &3.2361 & 6.7734    \\
%         Ours w/o incremental fusion & 26.72 & 0.9532 & 0.6359 &3.2361 & 6.7734    \\
%         Ours w/o \textit{FastPaint} & 27.82 & 0.9683 & 0.6574 & 3.7146 & 7.2136\\
%         \textit{WonderTurbo} & \textbf{28.65} &  \textbf{0.9732} & \textbf{0.6812} & \textbf{3.7253} & \textbf{7.3243} \\
        
%     \bottomrule
%     \end{tabular}
% \end{table}

\begin{table}[t]
    \centering
    \caption{Ablation study results on novel view renderings.}
    \vspace{-1.5mm}
    \footnotesize
    \label{tbl:ablation}
    \resizebox{\columnwidth}{!}{
    \begin{tabular}{lccccc}
    \toprule
         & CS$\uparrow$ & CC$\uparrow$ & CIQA$\uparrow$ & Q-Align$\uparrow$ & CA$\uparrow$ \\
        \midrule
        Ours w/o depth guided & 27.72 & 0.9532 & 0.6359 & 3.4361 & 7.1734    \\
        Ours w/o incremental fusion & 27.87 & 0.9654 & 0.6459 & 3.5431 & 7.2734    \\
        Ours w/o \textit{FastPaint} & 27.82 & 0.9683 & 0.6574 & 3.7146 & 7.2136\\
        \textit{WonderTurbo} & \textbf{28.65} &  \textbf{0.9732} & \textbf{0.6812} & \textbf{3.7253} & \textbf{7.3243} \\
    \bottomrule
    \end{tabular}
    }
    \vspace{-5.5mm}
\end{table}

\subsection{Ablation Study}

\noindent
\textbf{Geometry Modeling.} We compare different geometry modeling methods, including FreeSplat~\cite{wang2025freesplat} and DepthSplat~\cite{depthsplat}, all fine-tuned with the same settings for fairness. As shown in Tab. \ref{tbl:geometry}, FreeSplat \cite{wang2025freesplat} and DepthSplat~\cite{depthlab} underperform compared to \textit{StepSplat}, especially in Q-Align and CLIP aesthetic scores, due to their reliance on unsupervised depth estimation. In contrast, \textit{StepSplat} uses a consistent depth map to guide cost volume construction, enabling adaptive interactive 3D scene generation.

\noindent
\textbf{StepSplat.} We conduct ablation experiments on \textit{StepSplat} to illustrate the effect of the depth guided cost volume and the incremental infusion. As shown in Tab.~\ref{tbl:ablation}, the depth-guided cost volume is key to accurate geometry modeling and image quality. Meanwhile, incremental fusion improves performance by reducing redundant Gaussians and avoiding floating-point issues.

\noindent
\textbf{FastPaint.} We compare \textit{FastPaint} with the pretrained inpainting model~\cite{sd}. As shown in Tab.~\ref{tbl:ablation}, \textit{FastPaint} enhances the capability of 3D appearance modeling, with improvements across various metrics.

% \noindent
% \textbf{QuickFill.} To compare the results of using the interactive 3D generation dataset versus not using it. Our approach significantly outperform the baseline in terms of CLIP score and CLIP consistency, indicating that this dataset enhances the modeling of appearance.
\vspace{-0.3mm}
\section{Discussion and Conclusion}
\vspace{-0.3mm}
Despite progress in 3D scene generation from a single image, efficiency remains a challenge due to time-consuming geometry optimization and viewpoint refinement. To address this, we propose \textit{WonderTurbo}, an efficient framework for real-time interactive 3D scene generation that accelerates geometry optimization and appearance modeling. For accelerating geometry modeling, we introduce \textit{StepSplat}, which expands 3D scenes within 0.26 seconds while maintaining high visual quality, and \textit{QuickDepth}, which provides consistent depth priors for cost volume construction. For appearance modeling, \textit{FastPaint} is proposed to accomplish appearance modeling with only 2 inference steps while ensuring spatial consistency. Experiments show that \textit{WonderTurbo} accurately generates 3D scenes from text, outperforming CLIP-based metrics and user study win rates, with a 15$\times$ speedup.

 % user study We select four test cases, generating seven scenes for each, resulting in 28 scenes for comparison.  During each comparison, participants are shown two videos side-by-side, each produced by different methods following the same trajectory, with the order of the videos randomized to avoid bias. Participants then select the video they find most visually appealing.

% \begin{figure}[ht]
%     \centering
%     \includegraphics[width=0.8\textwidth]{figures/intro.pdf.pdf} % 假设文件名为 lu.pdf
%     \caption{This is a caption for the figure.}
%     \label{fig:my_label}
% \end{figure}

{
    \small
    \bibliographystyle{ieeenat_fullname}
    \bibliography{PaperForReview}
}
\newpage

\twocolumn[{%
\vspace{-1em}
\maketitle
\vspace{-2em}

\vfill % Pushes the content below to the bottom of the page

\begin{center}
\centering
\setlength{\abovecaptionskip}{0.5em}
\resizebox{0.99\linewidth}{!}{
\includegraphics{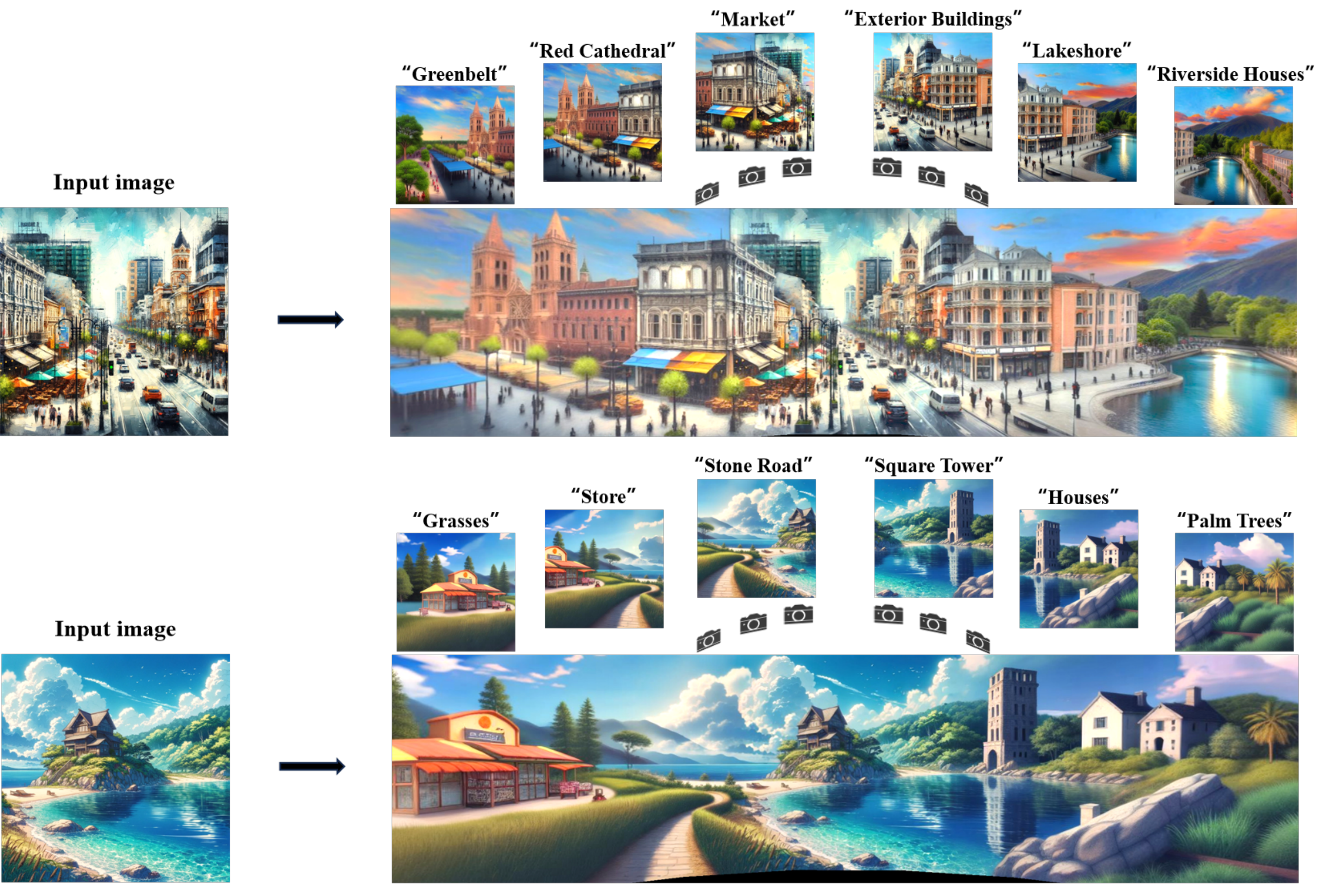}
}
\captionof{figure}{
 Qualitative examples.
}
\label{fig:123}
% \vspace{-0.5em}
\end{center}}]

\setcounter{section}{0}
\section{Implementation Details}
\noindent \textbf{Time cost evaluation.}  To effectively evaluate the generated scenes, we compare the time required to generate scenes of the same size. Specifically, for offline methods, we follow the setups of LucidDreamer~\cite{luciddreamer} and Text2Room~\cite{text2room}, first generating multiple images of new scenes and then converting them into 3D scenes according to their respective methods. For DreamScene360~\cite{dreamscene360} and Text2Room~\cite{text2room}, we use the Diffusion360~\cite{diffusion360} model to generate panorama images of the input images, which are then lifted to 3D. We compute the total time for this process and compute the time cost of generating the test scenes based on the size of the generated and test scenes. For online methods, we directly calculate the time required to generate a new scene.

\noindent \textbf{Metrics.} To evaluate the quality of 3D interactive scene generation, we utilize several metrics: CLIP scores (CS)~\cite{clip}, CLIP consistency (CC)~\cite{clip}, CLIP-IQA+ (CIQA)~\cite{clipiqa}, Q-Align~\cite{qalin}, and CLIP aesthetic scores (CA)~\cite{clip}. These metrics not only assess the quality of appearance modeling but also evaluate the quality of geometry modeling, as inaccurate geometry can lead to severe distortions when rendering novel views, which can significantly impact metrics such as CLIP-IQA+ (CIQA), Q-Align, and CLIP aesthetic scores (CA). The CLIP score (CS) measures the relevance between the scene prompt and the rendered image by computing the cosine similarity between their respective CLIP embeddings. CLIP consistency (CC) is assessed by measuring the cosine similarity between the CLIP embeddings of each novel view and the central view, ensuring semantic consistency across views. CLIP-IQA+ (CIQA) is an enhanced image quality metric that combines perceptual quality models with deep learning techniques to evaluate attributes. Finally, the CLIP aesthetic score (CA) captures the aesthetic quality of the image, considering elements like composition, contrast, and color harmony.

\section{Qualitative results}

\begin{figure*}[t] 
\centering
\includegraphics[width=6.5in]{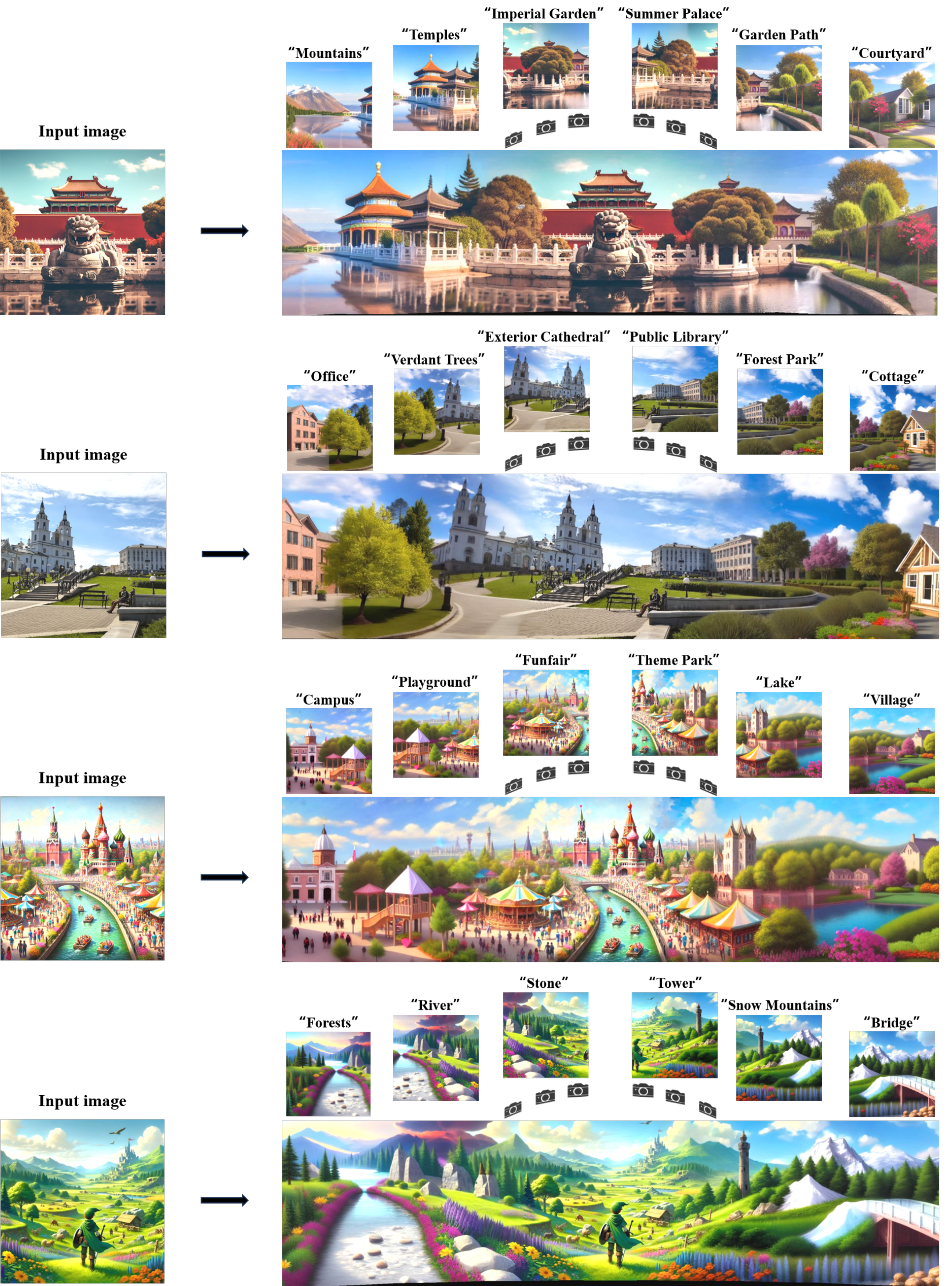}
\caption{Qualitative examples.}
\label{supply2}
\end{figure*}

\begin{figure*}[t!] 
\vspace{-30em}
\centering
\includegraphics[width=6.5in]{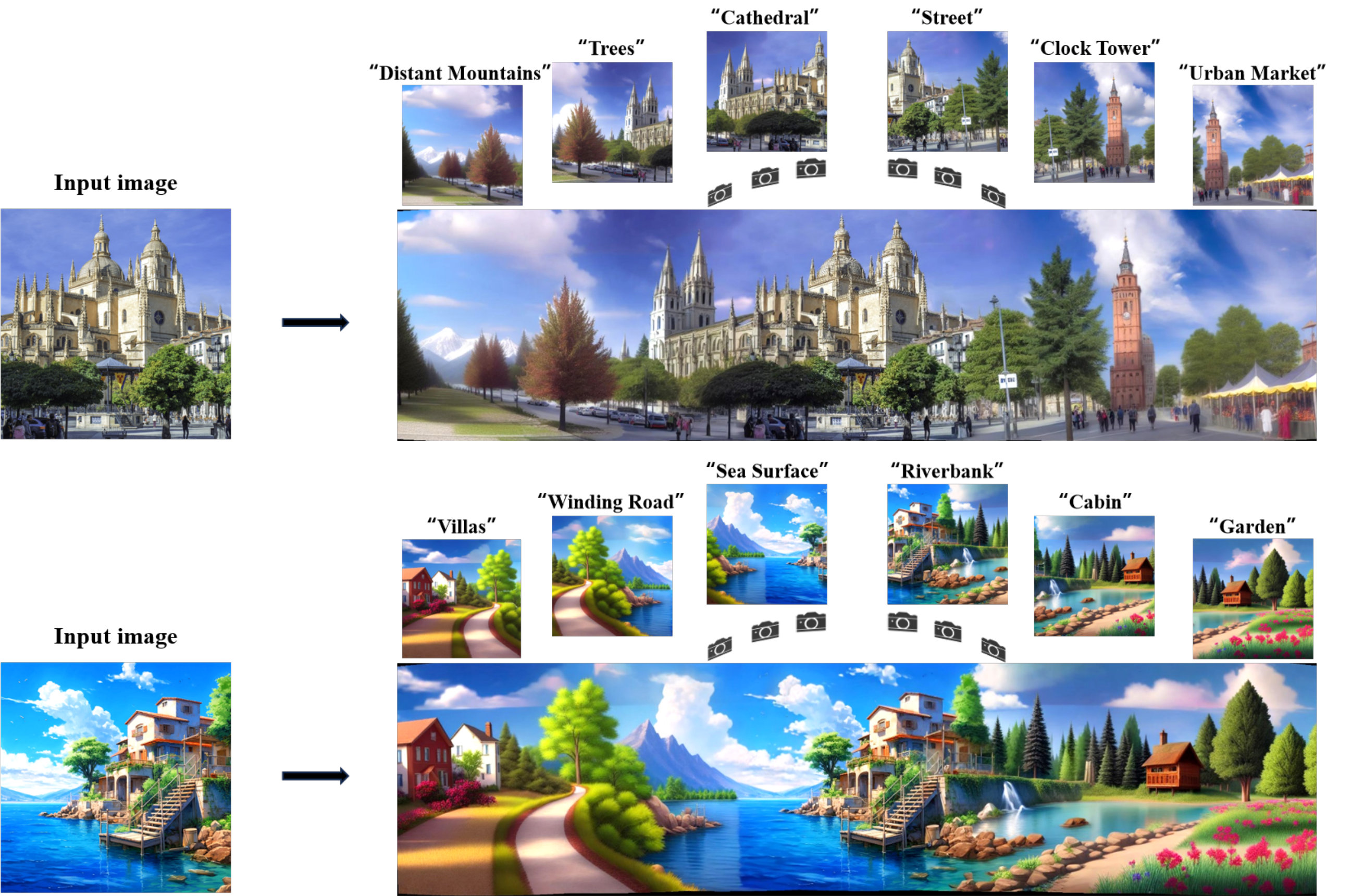}
\caption{Qualitative examples.}
\label{supply1}
\end{figure*}

As shown in Fig.~\ref{fig:123}, Fig.~\ref{supply1}, and Fig.~\ref{supply2}, we provide additional scenes in various styles to demonstrate the superiority of \textit{WonderTurbo}.

\end{document}